\documentclass{llncs}
\usepackage{url}
\usepackage{xspace}
\usepackage{color}
\usepackage{array}
\usepackage{underscore}
\usepackage{tipa}
\usepackage[T1]{fontenc}
\usepackage[utf8]{inputenc}
\usepackage{rotating}

\usepackage{amsmath}
\usepackage{amssymb}
\usepackage{booktabs}
\usepackage{graphicx}
\usepackage{listings}
\usepackage{paralist}
\usepackage{subfig}
\usepackage{hyperref}

\makeatletter

\def\holdocspecials{\do\ \do\$\do\&%
  \do\#\do\^\do\^^K\do\_\do\^^A\do\%}

\def\holtt{\trivlist \item[]\if@minipage\else\vskip\parskip\fi
\leftskip\@totalleftmargin\rightskip\z@
\parindent\z@\parfillskip\@flushglue\parskip\z@
\@tempswafalse \def\par{\if@tempswa\hbox{}\fi\@tempswatrue\@@par}
\obeylines \tt \let\do\@makeother \holdocspecials
 \frenchspacing\@vobeyspaces}

\makeatother

\newlength{\hsbw}
\newcommand\HOLSpacing{13pt}

   \newcommand\hilbert{\varepsilon}

   \newcommand{\Cond}{\(\rightarrow\)}
   \newcommand{\Eqv}{\(\equiv\)}
   \newcommand{\Iff}{\(\Longleftrightarrow\)\hspace{-1.5mm}}
   \newcommand{\Fa}{\(\forall\)}
   \newcommand{\Et}{\(\exists\)}
   \newcommand{\Eu}{\(\exists_{unique}\)}

   \newcommand{\Impl}{\(\Longrightarrow\)\hspace{-1.5mm}}
   \newcommand{\Func}{\(\to\)\hspace{-1.5mm}}

   \newcommand{\Lam}{\(\lambda\)}
   
   \newcommand{\Minus}{\(-\)}
   \newcommand{\Lminus}{\(-\)\hspace{-1.5mm}}
   \newcommand{\Prime}{\('\)}
   \newcommand{\Und}{\_}
   \newcommand{\Lt}{\(<\)}
   \newcommand{\Gt}{\(>\)}
   \newcommand{\Leq}{\(\leq\)}
   \newcommand{\Geq}{\(\geq\)}
   \newcommand{\Eq}{\(=\)}
   \newcommand{\Lrb}{\((\)}
   \newcommand{\Rrb}{\()\)}

   \newcommand{\Next}{\(\bigcirc\)}
   \newcommand{\Prev}{\(\ominus\)}
   \newcommand{\WPrev}{\(\widetilde{\bigcirc}\)}
   \newcommand{\Event}{\(\Diamond\)}
   \newcommand{\Once}{\(\underline{\Diamond}\)}  
\newcommand{\Hilbert}{\(\hilbert\)}

\newcommand{\Conj}{\(\wedge\)}
\newcommand{\Disj}{\(\vee\)}
\newcommand{\Neg}{\(\neg\)}
\newcommand{\Pnd}{\(\Diamond\)}

\newcommand{\Models}{\(\models\)}

\long\def\rechol#1#2#3{\let\next=\rechol\def\postnext{#2#3}\ifx#1\end
\let\next=\relax\def\postnext{\relax}
\else\ifx#1!\Fa                                          
\else\ifx#1@\Hilbert                                     
\else\ifx#1\#\Pnd                                        
\else\ifx#1'\Prime                                       
\else\ifx#1~\Neg                                         
\else\ifx#1\~\Neg
\else\ifx#1_\Und                                         
\else\ifx#1(\ifx#2+\ifx#3)\Next\def\postnext{}\fi        
            \else\ifx#2-\Prev\def\postnext{}             
            \else\ifx#2~\ifx#3)\WPrev\def\postnext{}\fi            
             \else\Lrb\fi\fi\fi                          
\else\ifx#1)\Rrb%
\else\ifx#1\/\Disj                                       
\else\ifx#1\.\Lam                                        
\else\ifx#1>\ifx#2=\Geq\def\postnext{#3}\else\Gt\fi      
\else\ifx#1?\ifx#2!\Eu\def\postnext{#3}\else\Et\fi       
\else\ifx#1-\ifx#2>\Func\def\postnext{#3}               
            \else\ifx#2-\Lminus\def\postnext{#3}
            \else\Minus\fi\fi                               
\else\ifx#1|\ifx#2-\Turns\def\postnext{#3}               
            \else\ifx#2=\Models\def\postnext{#3}
                 \else\Bar\fi\fi
\else\ifx#1<\ifx#2=\ifx#3>\Iff\def\postnext{}       
                   \else\Leq\def\postnext{#3}\fi    
            \else\ifx#2+\Event\def\postnext{}       
            \else\ifx#2-\Once\def\postnext{}       
            \else\Lt\fi\fi\fi                       
\else\ifx#1=\ifx#2=\ifx#3>\Impl\def\postnext{}            
                   \else\Eqv\def\postnext{#3}\fi         
            \else\ifx#2>\Cond\def\postnext{#3}
                 \else\Eq\fi\fi
\else\ifx#1/\ifx#2\^^M\Conj\par\def\postnext{#3}         
            \else\ifx#2\ \Conj\ \def\postnext{#3}\else#1\fi\fi  
\else#1\fi\fi\fi\fi\fi\fi\fi\fi\fi\fi\fi\fi\fi\fi\fi\fi\fi\fi\fi
\expandafter\next\postnext}

\newcolumntype{*}{>{\global\let\currentrowstyle\relax}}
\newcolumntype{^}{>{\currentrowstyle}}

\title{Developing Corpus-based Translation Methods between Informal and Formal Mathematics: Project Description\thanks{The final publication is available at http://link.springer.com.}}
\author{Cezary Kaliszyk \inst{1} \and Josef Urban \inst{2} \and Ji\v{r}\'i Vysko\v{c}il \inst{3}\thanks{Supported by the Grant Agency of Czech Republic Project GACR P103/12/1994.} \and Herman Geuvers \inst{2}}

\institute{University of Innsbruck, Austria \and
  Radboud University Nijmegen \and Czech Technical University}

\makeatletter
\renewcommand\section{\@startsection{section}{1}{\z@}%
                       {-12\p@ \@plus -4\p@ \@minus -4\p@}%
                       {8\p@ \@plus 4\p@ \@minus 4\p@}%
                       {\normalfont\large\bfseries\boldmath
                        \rightskip=\z@ \@plus 8em\pretolerance=10000 }}
\makeatother

\begin{document}
\maketitle
\vspace{-3mm}
\begin{abstract}
  The goal of this project\footnote{\url{http://mws.cs.ru.nl/~mptp/inf2formal}} is to (i) accumulate annotated
  informal/formal mathematical corpora suitable for training
  semi-automated translation between informal and formal mathematics
  by statistical machine-translation methods, (ii) to develop such
  methods oriented at the formalization task, and in particular (iii)
  to combine such methods with learning-assisted automated reasoning
  that will serve as a strong semantic component. We describe these
  ideas, the initial set of corpora, and some initial experiments done
  over them.
\end{abstract}

\vspace{-8mm}
\section{Introduction and Motivation Ideas}

Formal mathematics and automated reasoning are in some sense at the
top of the complexity ladder of today's precise
(``neat'') %
AI corpora and techniques. Many of us believe that practically all
mathematical theorems can be precisely formulated and that their
proofs can be written and verified formally, and that this carries
over to a lot of the knowledge accumulated by other exact
sciences. Given this unmatched expressivity and coverage, automated
reasoning over formal mathematics then amounts (or aspires) to being
the generic problem-solving technique for arbitrary problems that are
expressed in a sufficiently ``neat'' (formal) language and non-contradictory
setting.

The last ten years have brought significant progress in formalization
of mathematics and in automated reasoning methods for such formalized
corpora. Some graduate textbooks have been formalized, and we have produced general reasoning
methods that can often automatically find previous relevant knowledge
and prove many smaller steps and lemmas in such textbooks without the
necessity to manually provide any further hints or guidance.

However, even routine formalization is today still quite laborious,
and the uptake of formalization among mathematicians (and other
scientists) is very limited. There is a lot of cognitive processing
involved in formalization that is uncommon to majority of today's
mathematicians: formalization is a nontrivial skill to learn, and it
takes time. As a result, more than 100 years after Turing's birth,
most of mathematical (and scientific) knowledge is still largely
inaccessible to deep semantic computer processing.

We believe that this state of affairs can be today helped by
automatically \emph{learning} how to formalize (``semanticize'') informal texts, based on the knowledge available in existing
large formal corpora.%
There are several reasons for this belief:
\begin{enumerate}
\item 
Statistical machine learning (data-driven algorithm design) has been
responsible for a number of recent AI breakthroughs, such as web
search, query answering (IBM Watson), machine translation
(Google Translate), image recognition, autonomous car driving,
etc. As soon as there are enough data to learn from, data-driven
algorithms can automatically learn complicated sets of rules that
would be often hard to program and maintain manually.
\item With
the recent progress of formalization, reasonably large corpora are
emerging that can be (perhaps after additional annotation) used for
experiments with machine learning of formalization. The growth of such
corpora is only a matter of time, and automated formalization might
gradually ``bootstrap'' this process, making it faster and faster.
\item 
Statistical machine learning methods have already turned out
to be very useful in deductive AI domains such as automated reasoning
in large theories (ARLT), thus disproving conjectures that its
inherent undecidability makes mathematics into a special field where
data-driven techniques cannot apply.
\item Analogously, strong
semantic ARLT methods are likely to be useful in the formalization field also for
complementing the statistical methods that learn formalization.
This could lead to hybrid understanding/thinking AI
methods that self-improve on large annotated corpora by cycling
between (i) statistical prediction of the text disambiguation based on
learning from existing annotations and knowledge, and (ii) improving
such knowledge by confirming or rejecting the predictions by the
semantic ARLT methods. 
\end{enumerate}
The last point (4) is quite unique to the domain of (informal/formal)
mathematics, and a good independent reason to start with this AI
research. There is hardly any other domain where natural language
processing (NLP) could be related to such a firm and expressive
semantics as mathematics has, which is additionally to a reasonable
degree already checkable with existing ITP and ARLT systems. It is not
unimaginable that if we gradually manage to learn how mathematicians
(ab)use the normal imprecise vocabulary to convey ideas in the
semantically well-grounded mathematical world, such semantic grounding
of the natural mathematical language (or at least its underlying
mechanisms) will then be also helpful for better semantic treatment
of arbitrary natural language texts.

\vspace{-2mm}
\section{Approach}

The project is in the phase of preparing and analysing suitable
corpora, extracting interesting datasets from them on which learning
methods can be tried, collecting basic statistics about the corpora.
and testing initial learning approaches on them. Initially we consider the following corpora:

\noindent
\textbf{1. The various HOL Light developments:} in particular Flyspeck and
  Multivariate, for which we have a strong ARLT online service
  available~\cite{hhmcs}, and which is also in the case of Flyspeck
  and Multivariate aligned (by Hales) with the informal Flyspeck
  book. This is the main corpus we have so far worked on.  We have
  already written programs that collect the links between the informal
  and formal Flyspeck parts (theorems and definitions), and used such
  annotations for example for the joint informal/formal HTML
  presentation of Flyspeck~\cite{FlyspeckWiki}. Currently there are
  about 250-400 theorems mapped (using the \texttt{guid} tag defined
  by Hales), however we still need to improve our searching mechanism
  to find all the mapped informal/formal pairs in various parts of the
  library. In addition to the aligned theorems, Hales has also aligned
  over 200 concepts, which can be used as the ground level
  (dictionary) for the statistical translation algorithms. It is
  likely that further annotation of the texts will be useful, possibly
  also with some refactoring of the informal and formal parts so
  that they better correspond to each other. Most of the
  extraction/alignment chain is now automated so we can update our
  data after such transformations of the source texts.  We export the
  aligned theorems in several formats: parsed \LaTeX via \LaTeX{}ML
  (using libxml for querying), the original HOL text,
  bracketed HOL text suitable for parsing into external tools,
  internal (parsed and type-annotated) representation of the HOL
  theorems in a Lisp-like notation and in a XML notation, and also
  representation of each theorem in the (Prolog-parsable) THF TPTP
  format, containing type declarations of all constants recursively
  used by the theorems.
\\
\textbf{2. The Mizar/MML library:} and in particular its mapping to the book
  Compendium of Continuous Lattices~\cite{BancerekR02} (CCL) and a
  smaller mapping to Engelking's General Topology provided by
  Bancerek.\footnote{\url{http://fm.uwb.edu.pl/mmlquery/fillin.php?filledfilename=t.mqt&argument=number+1}}
  This is a potential large source of informal/formal pairs, however
  the MML has been developing quickly, and updating the mapping might
  be necessary to align the books with the current MML for which we
  have a strong online ARLT service~\cite{abs-1109-0616,KaliszykU13b}.
  We have also obtained the corresponding \LaTeX{} sources of the CCL
  book from Cambridge University Press, however we have not yet clarified the possible publication of the
  data extracted.%
\\
\textbf{3. The ProofWiki and PlanetMath informal corpora:} We have the XML
  and \LaTeX{} dumps of these wikis and have used them for initial
  experiments with disambiguation of informal texts in the student project
  \emph{Mathifier},\footnote{\url{http://mws.cs.ru.nl/~urban/Mathifier/}}
  motivated by the NLP work on Wikipedia
  disambiguation~\cite{RRDA11}. One relatively surprising preliminary
  result of this project is quite good performance (75\%) of the naive
  disambiguation algorithm using just the most frequent mathematical
  meaning without any additional context information. Another initial
  exploration was done on ProofWiki, whose relatively strict proof
  style is quite close to the Jaskowski-style natural deduction used
  in Mizar. We have measured this by mapping all math expressions
  and references in the ProofWiki sentences to just one generic
  expression/reference, and counted the frequency of various proof
  sentences. The
  results\footnote{\url{http://mizar.cs.ualberta.ca/~mptp/fpk1/opaqcounts1.txt}}
  again show great homogeneity of the corpus, where most of the proof
  discourse can be superficially mapped to Mizar natural deduction
  quite economically. Apart from defining and experimenting with such
  proof-level translation patterns, the main work on these corpora
  will be their mapping (possibly automated by using frequency
  analysis) to the Mizar and HOL Light corpora, in particular general
  topology that is developed a lot in ProofWiki and MML.
\vspace{-4mm}
\subsection{Methods, Tools and Planned Experiments}

There is a lot of relevant NLP research in (i) machine translation
(algorithms that directly translate between two languages) (ii)
word-sense disambiguation (algorithms that determine the exact meaning
of (sets of) words in sentences), and (iii) part-of-speech tagging and
phrasal and dependency parsing . The most successful statistical
methods (e.g., n-gram-based) require much larger corpora of aligned
data than we currently have, however some smarter algorithms such as
chart-parsing (the CYK) algorithm with probabilistic grammars (PCFGs)
should be usable already on the current scale of our data, perhaps
complemented by leaner memory-based approaches such as k-nearest
neighbor in the MBT toolkit.\footnote{\url{http://ilk.uvt.nl/mbt/}} Currently, we have started experimenting
with the Stanford parser,\footnote{\url{http://nlp.stanford.edu/software/lex-parser.shtml}} the Moses toolkit,\footnote{\url{http://www.statmt.org/moses/}} and our own Prolog/Perl
implementation of the (lexicalized) CYK algorithm on a subset of 500
formal (bracketed) Flyspeck expressions about trigonometric functions. 
Such initial experiments concern
relaxing of the precise disambiguated formal texts by adding more
ambiguity. For example whenever a casting functor (such as \texttt{Cx}
or \texttt{\&}) has to be used in the formal text, we can remove it,
and measure the success of the probabilistic parsing getting the
right formal meaning. Once such experiments produce good results, the
next step in this direction is learning the alignment of the
informal/formal text/trees using for example the tree-based learning
in the Moses toolkit.
The work with established tools such as the Stanford parser and
Moses will likely be complemented by our custom implementations that take advantage of the domain
knowledge. For example we can add immediate pruning of potential parse
trees in the CYK algorithm (or any chart parser) by using the HOL
Light (Hindley-Milner) type system or the Mizar (soft, dependent) type
system at each step of the algorithm.

\begin{small}
\vspace{-3mm}
\bibliographystyle{abbrv}
\bibliography{ate11}
\end{small}
\end{document}